\documentclass[12pt]{iopart}
\usepackage{graphicx}
\usepackage{caption}
\usepackage{fancyhdr}
\usepackage{booktabs}  
\usepackage{hyperref}
\usepackage{xcolor}
\usepackage{amsmath}
\usepackage{orcidlink}

\hypersetup{
  colorlinks=true,
  linkcolor=blue,
  citecolor=blue,
  urlcolor=green
}

\renewenvironment{equation*}{\begin{displaymath}}{\end{displaymath}}
\begin{document}

\title{A wearable Gait Assessment Method for Lumbar Disc Herniation Based on Adaptive Kalman Filtering }

\author{Yongsong Wang$^{1}$~\orcidlink{0009-0000-9545-4416}, Zhixin Li$^{2}$, Zhaohui Guo$^{3}$, Yin Ding$^{1}$, Zhan Huan$^{2,*}$ and Lin Chen$^{1}$~\orcidlink{0009-0008-6486-7988}}

\address{$^{1}$ School of Computer Science and Artificial Intelligence, Changzhou University, Changzhou,Jiangsu 213000, China }
\address{$^{2}$ School of Microelectronics and Control Engineering, Changzhou University, Changzhou, Jiangsu 213000, China}
\address{$^{3}$ Department of Orthopedics, The Affiliated Changzhou No.2 People’s Hospital of Nanjing Medical University, 29 Xinglong Alley, Changzhou, China}
\address{$^{*}$ Author to whom any correspondence should be addressed.}

\ead{hzh@cczu.edu.cn}
\vspace{10pt}

\begin{indented}
  \item[]{December 2023} \par
\end{indented}

\begin{abstract}
Lumbar disc herniation (LDH) is a prevalent orthopedic condition in clinical practice. Inertial measurement unit sensors (IMUs) are an effective tool for monitoring and assessing gait impairment in patients with lumbar disc herniation (LDH). However, the current gait assessment of LDH focuses solely on single-source acceleration signal data, without considering the diversity of sensor data. It also overlooks the individual differences in motor function deterioration between the healthy and affected lower limbs in patients with LDH. To address this issue, we developed an LDH gait feature model that relies on multi-source adaptive Kalman data fusion of acceleration and angular velocity. We utilized an adaptive Kalman data fusion algorithm for acceleration and angular velocity to estimate the attitude angle and segment the gait phase. Two Inertial Measurement Units (IMUs) were used to analyze the gait characteristics of patients with lumbar disc issues and healthy individuals. This analysis included 12 gait characteristics, such as gait spatiotemporal parameters, kinematic parameters, and expansibility index numbers. Statistical methods were employed to analyze the characteristic model and confirm the biological differences between the healthy affected side of LDH and healthy subjects. Finally, a classifier based on feature engineering was utilized to classify the gait patterns of the affected side of patients with lumbar  disc disease and healthy subjects. This approach achieved a classification accuracy of 95.50\%, enhancing the recognition of LDH and healthy gait patterns. It also provided effective gait feature sets and methods for assessing LDH clinically.
\end{abstract}

\section{Introduction}

Lumbar disc herniation (LDH) is a common musculoskeletal disorder that affects approximately 5\% of the population. It causes nerve root irritation either mechanically or through inflammatory mediators and results in radiating pain, called sciatica \cite{Alakokko2002GeneticRF}. Pain is an important indicator for preoperative and postoperative evaluation of patients with clinical and imaging manifestations of lumbar disc herniation. Radiating leg pain can lead to unilateral or even bilateral numbness of the lower limbs. LDH affects gait kinematics, resulting in three-dimensional movements of LDH patients obstacles, and a lack of synergy \cite{Kuligowski2021LumbopelvicBI}. Some studies evaluated gait parameters from different gait patterns by using medical laboratory-based platforms, Lee et al. used a gait photovoltaic system to evaluate the relationship between pain and gait instability in patients with lumbar disc herniation (LDH)\cite{3Lee2021AssociationBP}. Masoud et al. used the WIN-TRACK gait analysis platform to study the relationship between LDH complicated by chronic mechanical low back pain (CMLBP) and pain and verified the important clinical symptoms of gait abnormalities caused by sciatic nerve pain in LDH patients \cite{4AmirRashediBonab2019AssessmentOS}. Chen et al. constructed a conditional deep convolutional generative adversarial network (CDCGAN) based on MRI feature extraction, three-dimensional modeling, and CDCGAN model classification to assist in the diagnosis of LDH, and combined MRI and LDH medical features to classify LDH \cite{5Chen2021UsageOI}. The successful application of medical laboratory platforms has promoted clinical movement and posture assessment of LDH, but the high cost is not suitable for public use.

Inertial motion unit sensor (IMUs) monitoring of LDH impaired gait is a low-cost and portable way. IMU acceleration and angular velocity data are often used to estimate the spatiotemporal parameters of gait to characterize impaired and healthy gait. By analyzing different gaits, pattern analysis to reveal the recovery status and pathological characteristics of patients with motor and neurological diseases \cite{6Trojaniello2014EstimationOS,7Mantashloo2023LowerBK}. Many studies have calculated gait spatiotemporal parameters and joint kinematic parameters through IMU to quantify abnormal gait and movement disorders in diseases, such as stroke, Parkinson's disease, chronic mechanical low back pain, etc\cite{8,9,10}. There are currently relatively few studies using IMU equipment to evaluate LDH gait patterns. Existing LDH gait parameter models focus on the processing of gait spatiotemporal parameters. These parameters reflect the gait differences and movement abilities of the healthy and affected side of LDH in the spatiotemporal dimension, but the kinematic parameters and expansibility indexes are ignored, making it difficult to characterize the overall gait ability and movement posture \cite{11,12,13,14}; however, kinematic parameters and expansibility indexes such as knee joint motion angle and variability have been successfully applied to different disease stages. In the evaluation of dynamic models \cite{15,16,17}. LDH has unique pathological gait characteristics. Exploring more advanced kinematic parameters and expanded indicators to indicate the gait deterioration pattern of LDH patients is very important for analyzing the reasons for the loss of lower limb gait rhythm caused by lumbar degenerative disease \cite{18}.

In previous IMU gait classification tasks, acceleration and angular velocity data have been used as input for deep learning to identify abnormalities in different gait impairment patterns. Tan et al. used acceleration data to detect heel strike (HS) and toe off (TO) in the user's gait cycle in an improved long short-term memory (LSTM) network to identify different gaits \cite{19}. Dmitry et al. proposed a deep learning model for detecting gait events based on IMU acceleration and angular velocity, and analyzed the effectiveness of various pathological gait patterns \cite{20}, but the input to the convolutional layer of the model is single-modality data. Djordje et al. facilitated the automatic classification of gait disorders by performing machine learning classification on ground reaction forces of a large number of healthy and pathological gait samples, and revealed that data scale imbalance will ultimately affect the classification results \cite{21}. These methods provide rich and effective means for identifying abnormal gait in diseases such as LDH, but deep learning ignores important kinematic information, making it difficult to interpret complex pathological structures. In addition to these end-to-end disease gait automatic classification and diagnosis models, it is of great significance to explore reliable biological feature sets to characterize the lower limb motor function of LDH patients to assist physicians in clinical assessment of the disease.

The fusion of multi-source data from IMU in gait analysis can improve the comprehensiveness of gait injury assessment \cite{22,23}. Multi-source data fusion is used in pedestrian navigation systems to reproduce movement postures by utilizing a variety of data fusion algorithms. With mature applications \cite{24,25}, Lin et al. established a movement assessment framework based on the fusion of surface electromyography (sEMG) and IMU features to evaluate the gait difference before and after recovery in patients with acute stroke \cite{26}. Dovin inputs the weighted accumulation of acceleration and angular velocity into an unsupervised gait event recognition method to quantify the gait differences of runners at different speeds \cite{27}. Sudsanguan et al. combined knee motion angle and foot pressure data to analyze changes in individual motion patterns \cite{28}. Qiu et al. designed a multi-sensor data fusion algorithm based on complementary filtering and error correction of gait parameters and analyzed the fluctuation of joint angles and the asymmetry of foot elevation during walking \cite{29}. Guo et al. proposed a new method of transferable multi-modal fusion using electromyography and gyroscope data to continuously predict knee joint angles and corresponding gait phases by fusing multi-modal signals \cite{30}. Previous gait assessment models focused on the weighting and accumulation of multiple types of raw data. They did not perform kinematic modeling of human posture on IMU acceleration and angular velocity data and did not achieve the ideal fusion effect.

In this study, because of the lack of consideration of angular velocity information and the limitations of spatiotemporal parameters in the current LDH gait assessment model, a gait parameter assessment model for the healthy and affected side of LDH was constructed using multi-source data based on acceleration and angular velocity of two IMUs attached to the calf, including gait characteristics in three gait modes. A gait analysis model based on the adaptive Kalman multi-source data fusion algorithm is proposed. The acceleration and angular velocity data are fused to obtain a gait pattern in the form of an attitude angle to segment the gait phase, and the Eckmann algorithm is used to convert the motion attitude into a chaotic system. The maximum Lyapunov index is calculated to evaluate the stability of the motion state, and multi-source and multi-type gait features such as gait spatiotemporal parameters, expansibility indexes, and kinematic parameters are extracted. Finally, a machine learning classifier based on feature engineering was used to identify LDH gait abnormalities and was more effective than previous studies in classifying LDH gait, and a gait feature set was obtained to assist in the clinical diagnosis of LDH.

This paper has been organized as follows. Construction of LDH gait data set are presented in section\ref{sec2}. section\ref{sec3} details the methodology. Results scheme and results
are explained in section\ref{sec4} and we discuss the results
in section\ref{sec5}. Conclusions of the entire research have
been given in section \ref{sec6}.

\section{Construction of LDH gait data set}\label{sec2}
\subsection{\textbf{Participants}}
\begin{figure}
\centering
  \includegraphics[width=0.6\textwidth]{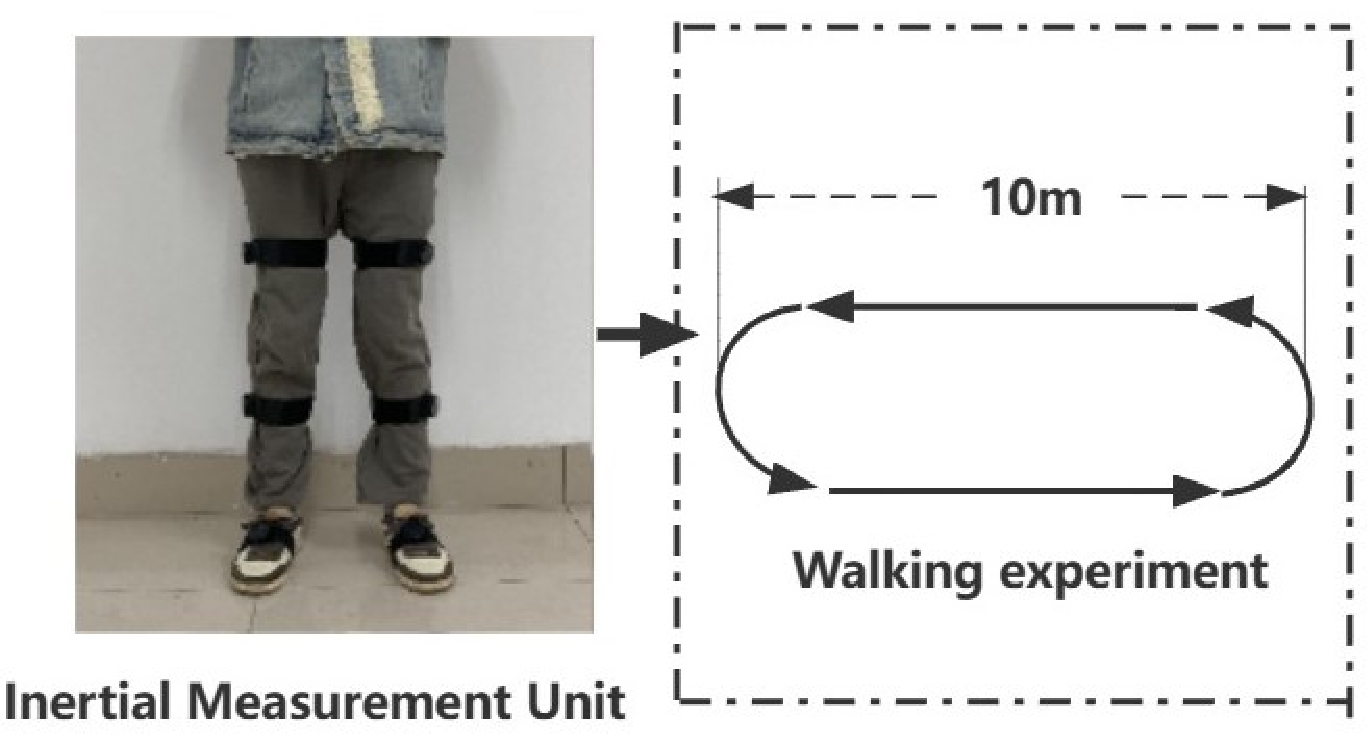}
  \caption{Walking experiment diagram.}
  \label{fig1}
\end{figure}

This study recruited 20 lumbar disc patients and 15 healthy subjects. The demographic data of LDH participants are shown in Table\ref{tab1}. All LDH subjects presented with a certain degree of radiating leg pain and low back pain and the symptoms only occurred on one side. They were clinically diagnosed with lumbar disc herniation and were eligible for surgery. The consent of all subjects was obtained before surgery. Patients with infection, cancer, previous lumbar spine surgery, and other pathologies that could alter gait, including knee, hip, or neurological dysfunction, were excluded. As shown in figure\ref{fig1}, each LDH patient underwent three walking experiments before surgery. In each experiment, the subjects were asked to walk back and forth in the most natural way on flat ground without auxiliary support. At about 20 meters, the sensor records acceleration, gyroscope, and other data during walking and stores it in a CSV file. Healthy subjects were also recruited as a control group, following a similar process. All participants showed normal gait patterns and had no previous history of LDH and gait disorders. Each healthy subject was asked to Walk most naturally on flat ground for about 10 meters.

\begin{table}[htbp]%调节图片位置，h：浮动；t：顶部；b:底部；p：当前位置
	\centering
	\caption{Demographic data collected from participants recruited for the study}
	\label{tab1}  
	\begin{tabular}{ccc}%表格中的数据居中，c的个数为表格的列数
		\hline\hline\noalign{\smallskip}	
		Features & LDH subjects & Healthy subjects\\
		\noalign{\smallskip}\hline\noalign{\smallskip}
		Age mean($\pm$ SD) & 52($\pm$20) & 24($\pm$3) \\
		Height(cm), mean($\pm$SD) & 167($\pm$12) & 173($\pm$10)  \\
            Weight(kg), mean($\pm$SD) & 65($\pm$10) & 67($\pm$19)  \\
            Disease level &  &   \\
            VAS & 5.8 &  0.5 \\
            ODI & 44.1 &  1.2 \\ 
            JOA & 19.2 &  29 \\
            Male & 11 & 11  \\ 
            Female & 6 & 4  \\ 
		\noalign{\smallskip}\hline
	\end{tabular}
\end{table}
\subsection{\textbf{Data acquisition}}
This study uses two IMUs, including a three-dimensional accelerometer with a range of ±8g and a three-dimensional gyroscope with a range of ±1000°/s for gait analysis. Each IMU is attached near the ankle joint of each calf and placed in the lateral sagittal plane. The x-axis of each IMU is along the handle in the sagittal plane; the y-axis in the sagittal plane is in the anterior-posterior direction of the handle; the z-axis is in The middle and lateral sides of the handle are perpendicular to the sagittal plane \cite{31}. For potential errors caused by accelerometer and angular velocity meter drift, each sensor was verified using the optical Vicon Nexus system (Vicon Motion Systems Ltd, Yarnton, UK) provided by Sebastian Glowinski \cite{32}, which we treated as as the gold standard.  Each IMU is sampled at 100 Hz and the measurements are saved to a database csv file. The experiment with the most stable walking status among the three experiments was selected for analysis. The acceleration data of the left and right calves were independently analyzed based on the results of clinical diagnosis of the healthy and affected sides. In order to improve the calculation accuracy, the data after turning and turning were removed and analyzed only once. Experiment with the process of one-way walking. An algorithm was implemented on Python and a database was established to store and analyze the collected data.

\subsection{\textbf{Ethical support}}
This study was performed in accordance with the Declaration of Helsinki. This human study was approved by the Clinical Research Committee of Changzhou Second People’s Hospital - approval: [2022]KY306-01. All adult participants provided written informed consent to participate in this study.

\subsection{\textbf{Data set}}
The data set is divided into two types of samples: LDH patients and healthy subjects. The acceleration, angular velocity and gyroscope data collected by the IMU are stored in csv files. Each IMU is collected separately and the wearing position is unified. The cut-off frequency during the collection process is 100hz. Acceleration and gyroscope data during walking are intercepted from the csv file and stored in a new csv file, and the time axis remains unchanged. The data set contains left and right leg acceleration and gyroscope data of 20 LDH patients and 15 healthy subjects, with a total of 70 csv files.

\section{\textbf{Method}}\label{sec3}

\begin{figure*}[htpb]
    \centering
    \includegraphics[width=\textwidth,height=0.5\textheight,keepaspectratio]{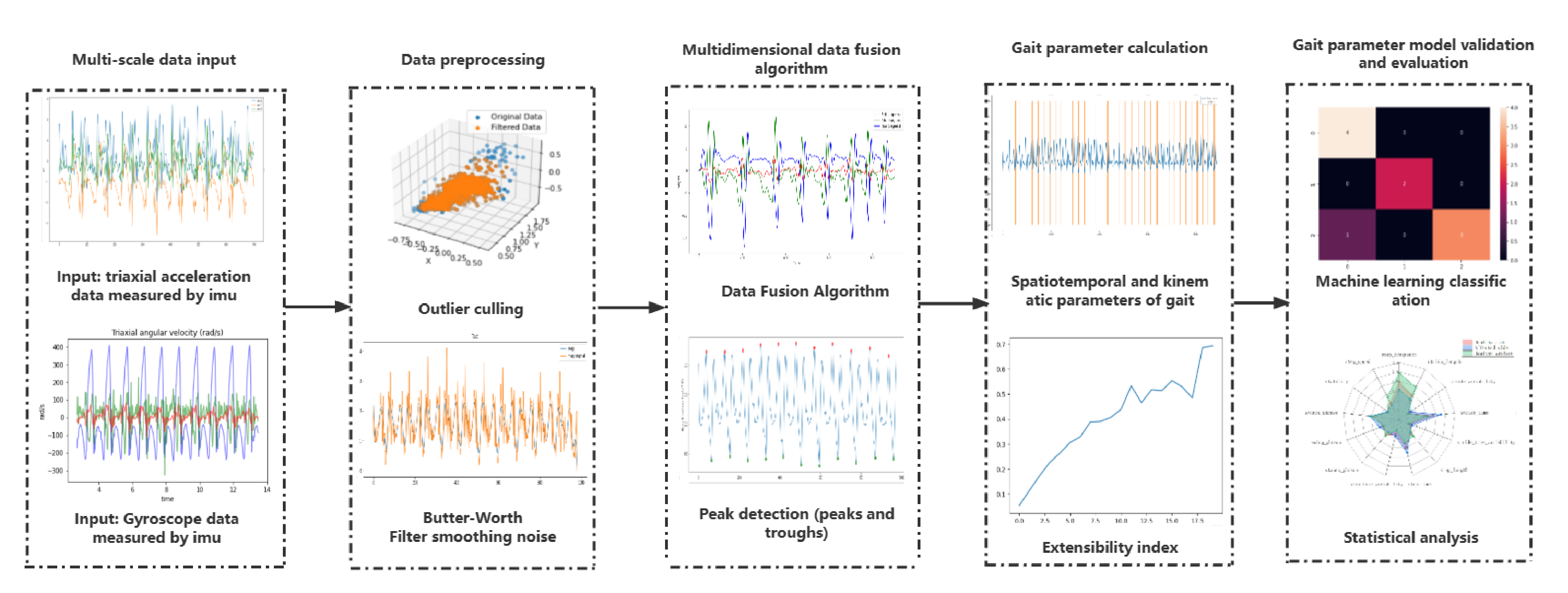}
    \caption{Flow chart of lumbar disc herniation gait diagnosis model based on IMU multi-source data fusion.} 
    \label{fig2}
\end{figure*}

Figure\ref{fig2} is a flow chart of the gait diagnosis model for lumbar disc herniation. First, data preprocessing operations are performed on multi-source data from accelerometers and gyroscopes, including outlier removal and smoothing denoising; then the smoothed acceleration and angular velocity are fused to calculate attitude angles and divide gait phases; using a peak detection algorithm to detect wave peaks and troughs to obtain heel strike and toe off events. Gait spatiotemporal parameters such as step length and gait cycle are calculated from gait events; at the same time, motion posture data are obtained to evaluate the stability and stability of gait respectively. Predictability; to verify the effectiveness of the gait parameter model, we used statistical analysis and feature engineering-based classifiers to analyze the biological differences and classification effects between the affected side of LDH and healthy subjects. The specific method will be introduced in detail below.

\subsection{\textbf{Data preprocessing}}

\begin{figure*}
\centering
  \includegraphics[width=0.7\linewidth]{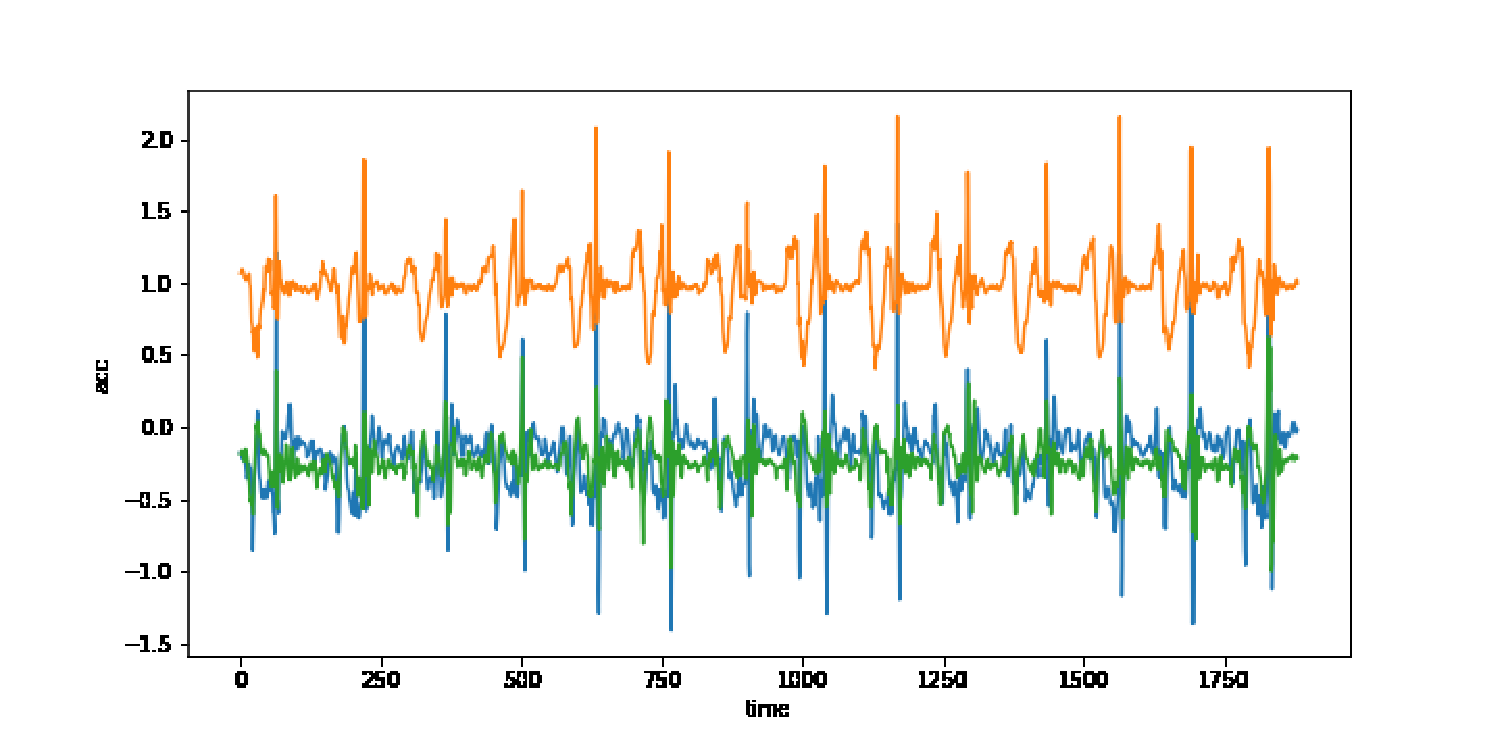}   
  \includegraphics[width=0.7\linewidth]{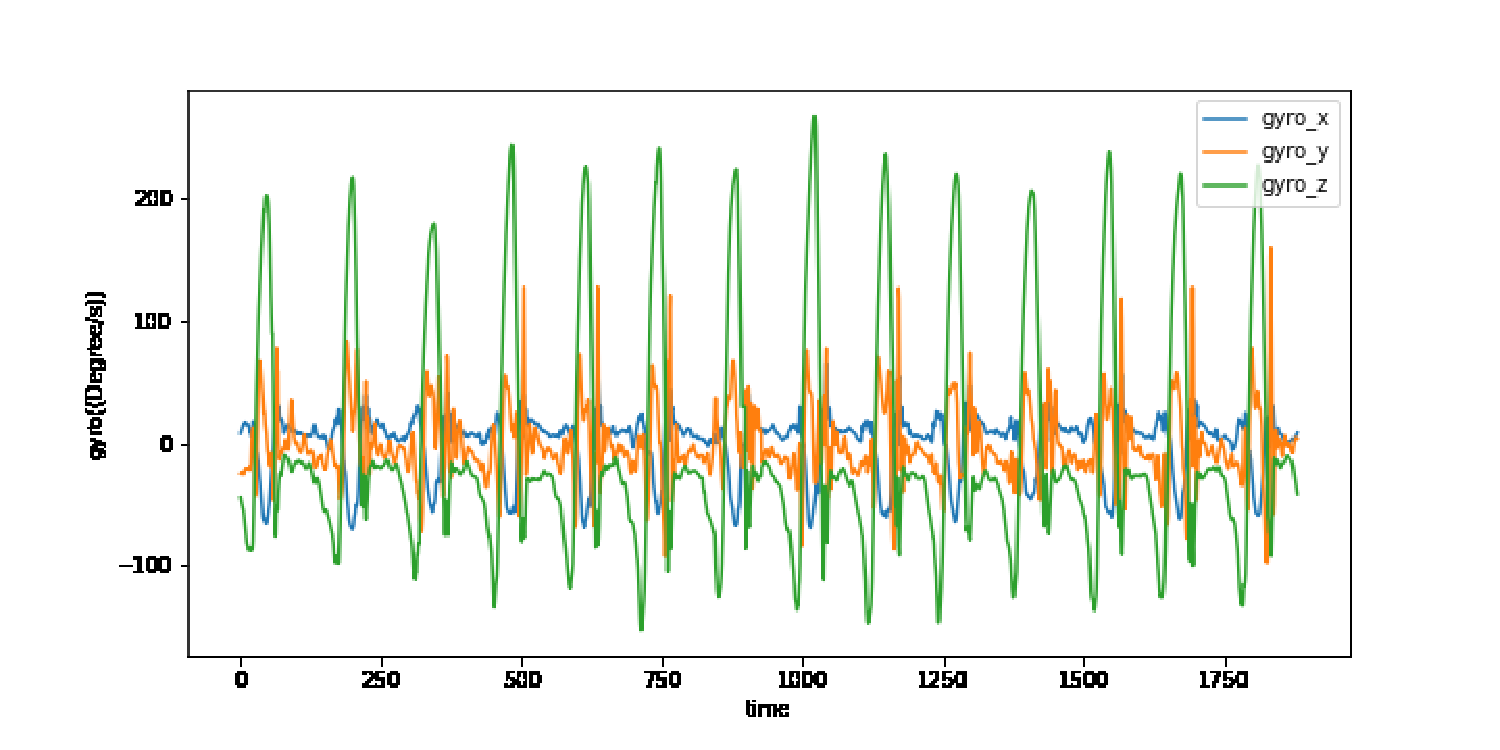}
  \caption{Preprocessed initial acceleration and angular velocity data.}
  \label{fig3}
\end{figure*}

Data preprocessing is performed on the acceleration and angular velocity data from the unilateral lower leg. The acceleration is first synthesized as shown in 
equation\ref{eq1}.
\begin{equation}acc={\sqrt{accx^{2}+accy^{2}+accz^{2}}}.\label{eq1}\end{equation}
Where accx, accy and accz are x, y and z axis accelerations respectively. Since it was verified in \cite{16} that the turning process can improve the recognition accuracy of ill-conditioned gait, we intercepted the data of the linear walking process to calculate the signal amplitude and eliminate the abnormal values according to the standard deviation, with a threshold value of three times the standard deviation. The purpose is to reduce the measurement error caused by different wearing positions of IMUs. As shown in figure\ref{fig3}, the acceleration and angular velocity data are subjected to a fourth-order Butterworth low-pass filter with a cutoff frequency of 10hz to smooth high-frequency noise. Then, the filtered data is interpolated to reduce the error caused by signal interference during the IMU data collection process, and finally obtain the preprocessed data.

\subsection{\textbf{Gait solution model based on adaptive Kalman data fusion}}
Accelerometers and gyroscopes are sensors that measure speed state variables during walking.Errors produced by the sensors during the measurement process are unavoidable, and the gyroscope will drift during long-term measurement processes. Since the accelerometer and gyroscope are independent in the measurement process, the adaptive Kalman filter algorithm can reduce the accompanying errors by fusing the acceleration and angular velocity data \cite{33,34}. We fuse acceleration and angular velocity data to obtain the attitude angle of the motion process, which can adapt to the changing environment and takes into account noise and uncertainty to provide accurate state estimation. The state space model of the adaptive Kalman filter includes a state transition equation and an observation equation. The state equation is used to describe the change process of the subject's motion state over time, and the observation equation maps the state vector at the current moment to the observation space.

Due to the difference between the IMU coordinate system and the earth coordinate system, to solve the subject's ground attitude, it is first necessary to convert the gyroscope data [p, q, r] in the IMU coordinate system to the Northeastern Sky (ENU) coordinate system:
\begin{equation}\begin{bmatrix}\text{gyrox}\\ \text{gyroy}\\ \text{gyroz}\end{bmatrix}=R\cdot\begin{bmatrix}p\\ q\\ r\end{bmatrix}\end{equation}

\begin{equation}
R = \begin{bmatrix}
    \text{c}(\theta)\text{c}(\psi) & -\text{c}(\phi)\text{s}(\psi) + \text{s}(\phi)\text{s}(\theta)\text{c}(\psi) & \text{s}(\phi)\text{s}(\psi) + \text{c}(\phi)\text{s}(\theta)\text{c}(\psi) \\
    \text{c}(\theta)\text{s}(\psi) & \text{c}(\phi)\text{c}(\psi) + \text{s}(\phi)\text{s}(\theta)\text{s}(\psi) & -\text{s}(\phi)\text{c}(\psi) + \text{c}(\phi)\text{s}(\theta)\text{s}(\psi) \\
    -\text{s}(\theta) & \text{s}(\phi)\text{c}(\theta) & \text{c}(\phi)\text{c}(\theta)
\end{bmatrix}
\end{equation}

Where R is the direction matrix; $\phi,\theta,\psi $ are the initial attitude angles; c and s are cos and sin trigonometric functions respectively, gryox, gyroy, and gyroz represent the three-axis acceleration values based on the earth coordinate system.

The adaptive Kalman filter state model is:
\begin{equation}X_{k}=AX_{k-1}+Bu_{k-1}+W_{k}\end{equation}

\begin{equation}A=\begin{bmatrix}1&0&0\\0&1&0\\0&0&1\end{bmatrix}, B=\begin{bmatrix}\Delta t&0&0\\0&\Delta t&0\\0&0&\Delta t\end{bmatrix}\end{equation}

Where A is the state transition matrix, B is the input matrix, and $W_{k}$ is the process noise vector; $X_{k-1}$ is the 3-dimensional error state vector containing the Euler angle value calculated with angular velocity.

\begin{equation}\left.X_{k-1}=\left[\begin{array}{ccc}{\dot{\phi},}&{\dot{\theta},}&{\dot{\psi}}\\\end{array}\right.\right]\end{equation}

\begin{equation}\begin{cases}\dot{\phi}=\text{gyrox}+\left(\frac{\sin(\theta)\sin(\phi)}{\cos(\theta)}\right)\cdot\text{gyroy}+\left(\frac{\sin(\theta)\cos(\phi)}{\cos(\theta)}\right)\cdot\text{gyroz}\\\dot{\theta}=\cos(\phi)\cdot\text{gyroy}-\sin(\phi)\cdot\text{gyroz}\\\dot{\psi}=\frac{\sin(\phi)}{\cos(\theta)}\cdot\text{gyroy}+\frac{\cos(\phi)}{\cos(\theta)}\cdot\text{gyroz}\\\end{cases}\end{equation}

\begin{equation}u_{k-1}=\begin{bmatrix}\text{gyrox}\\\text{gyroy}\\\text{gyroz}\end{bmatrix}\end{equation}

Where $\dot{\phi}$, $\dot{\theta}$ and $\dot{\psi}$ are respectively expressed as the derivatives of attitude Euler angle ${\phi}$, ${\theta}$ and ${\psi}$ with respect to time.

The observation model :
\begin{equation}Z_{k}=HX_{k}+V_{k}\end{equation}

Where $Z_k$ is the attitude angle calculated from the acceleration as the observed value at time t, H is the observation matrix; $V_{k}$ is the measurement noise, assumed to be Gaussian white noise, and its detailed form is as follows:

\begin{equation}\left.Z_k=\left[\begin{array}{c}roll_{acc}\\pitch_{acc}\\yaw_{acc}\end{array}\right.\right]=\left[\begin{array}{ccc}1&0&0\\0&1&0\\0&0&1\end{array}\right]X_{k}+\left[\begin{array}{c}N_{roll}\\N_{pitch}\\N_{yaw}\end{array}\right]\end{equation}

Where $N_{roll}$, $N_{pitch}$ and $N_{yaw}$ are respectively the observation noise of the three-dimensional attitude angle; $roll_{acc}$, $pitch_{acc}$ and $yaw_{acc}$ represent the attitude Euler angle calculated from the acceleration. Adaptive Kalman filtering to fuse acceleration and angular velocity includes two steps: prediction and update \cite{25}:

Prediction step: first we convert the state equation and observation equation into discrete form,
\begin{equation}\begin{cases}Z_{k}=HX_{k}+N,\\X_{k}=F_{k/k-1}X_{k-1}+W_{k-1},\end{cases}\end{equation}

Status prediction:
\begin{equation}\hat{\boldsymbol{X}}_{k/k-1}=F_{k/k-1}\hat{\boldsymbol{X}}_{k-1}\end{equation}

Update steps: status update and Kalman gain,
\begin{equation}\begin{aligned}K_{k}&=P_{k/k-1}H_{k}^T\left(H_{k}P_{k/k-1}H_{k}^T+R_{k}\right)^{-1}\\\widehat{X}_k&=\widehat{X}_{k/k-1}+K_k(z_k-H_k\widehat{X}_{k/k-1}),\end{aligned}\end{equation}

Forecast and estimated mean square errors:
\begin{equation}\begin{cases}P_{k/k-1}=F_{k/k-1}P_{k-1}F^T_{k/k-1}+Q_{k-1},\\P_{k}=(I-K_{k}H_{k})P_{k/k-1}(I-K_{k}H_{k})^T+K_{k}R_{k}K^T_{k},\end{cases}\end{equation}

Where $P_{k}$ is the state covariance matrix representing the uncertainty of state estimation, Q is the process noise covariance matrix, and the noise statistical characteristics of the subject's motion state are estimated based on the actual angular velocity observation data, and the Kalman filter is dynamically adjusted. Q and the observation noise covariance matrix R.

The adaptation process is to assign weights $(\beta_{i},i=1,2,...,N)$ according to the total number of constructed Kalman filters (N),  and each estimation of the motion state system is based on a different outlier detection threshold:

\begin{equation}\hat{x}_{k}=\sum_{i=1}^{L}\hat{x}_{k}^{i}\beta_{i},\quad i=1,2,\ldots,L\end{equation}

\begin{figure}
\centering
  \includegraphics[width=0.6\textwidth]{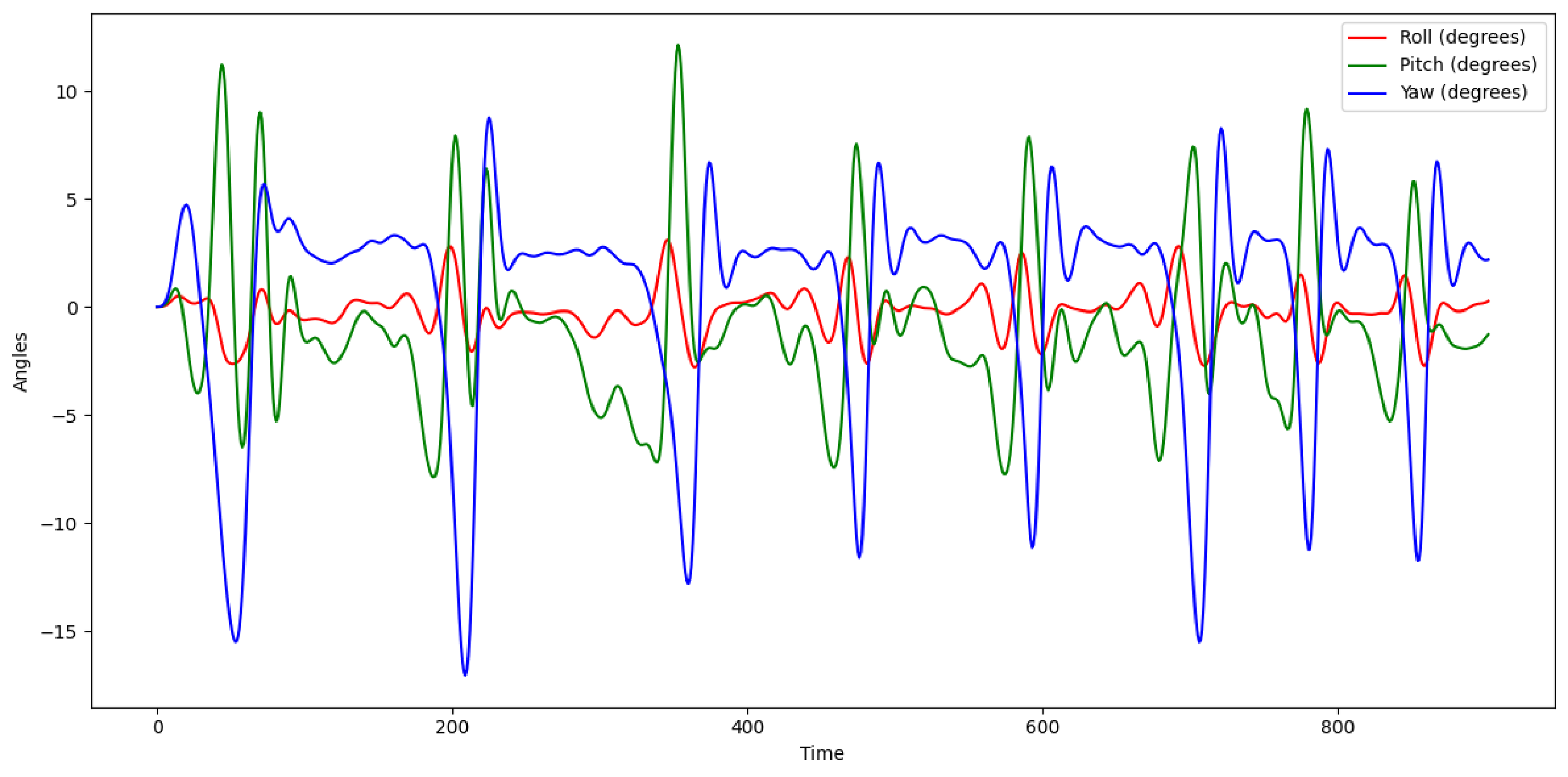}
  \caption{Three-dimensional attitude angle estimated by adaptive Kalman fusion algorithm.}
  \label{fig4}
\end{figure}

Where $\hat{x}_{k}$ is the estimated value of the system and $\hat{x}_{k}^{i}$ is each state estimate.The estimation error is calculated by comparing the observed values and the predicted observed values, and the Kalman gain is used to correct the predicted state vector and state covariance matrix to obtain an accurate estimate of the attitude Euler angle. The results are shown in figure\ref{fig4}. The pitch angle pitch data was extracted for peak detection to obtain the peaks and troughs to divide the gait phase into the standing phase and the swing phase. The stride frequency, stride time, and single step were obtained through the method provided by Alexander Rampp et al. calculation of time\cite{35}, in which stride length , step length and pace are estimated by fusing three-axis acceleration. The three-dimensional attitude angle obtained by fusion has a more accurate estimate. According to the pitch angle and roll angle after coordinate system conversion, the knee joint bending angle and swing angle are estimated by cumulative integration; the parameter definitions are shown in table\ref{tab2}.

\begin{table}
    \caption[Short Caption for List of Figures]{Definition of gait spatiotemporal parameters.}
    \label{tab2}
        \setlength{\tabcolsep}{3pt}
\footnotesize
\begin{tabular}{p{120pt}p{310pt}}
\br
 Parameters&Definition\\
\mr
Step frequency(SF)& The number of steps per second.\\
Step length(SL)& The distance from the heel of one foot to the heel of the foot again. \\
Step speed(SS)& Overall straight-line distance traveled in the direction of travel per unit of time. \\
Stride time(ST)& The duration between two adjacent heel landing events. \\
Stance phase(STP)&  The ratio of the duration between HS and TO events in gait cycle \\
Swing phase(SWP)& The ratio of the duration between adjacent TS and HS events during  gait cycle. \\
Step time(STT) & The average time from heel landing on one side to heel landing on the other. \\
WQK & Maximum angle of flexion after lower limb extension into the early stage of STP.\\
BDK & Maximum angle of swing after lower limb extension into the early stage of STP.\\
\br
\end{tabular}\\
$^{a}$HS:heel-strike events;TS:toe-strike; TO:toe-off; $^{b}$WQK:Maximum knee bend Angle; BDK:Maximum knee joint Swing Angle
\end{table}
\normalsize

\subsection{\textbf{Gait extensibility index}}
\begin{enumerate}
\item Variability (V) : calculated as the ratio of the standard deviation of the gait parameter to the mean value. The higher the variability index, the worse the gait performance.

\begin{equation}variability = \frac{sd}{mean}\end{equation}

\item Gait Stability (STA):the Eckmann algorithm is used to calculate the maximum Lyapunov index to evaluate the stability index. Its basic principle is based on the phase space reconstruction method, which can convert the original time series data into the trajectory in the high-dimensional phase space, so as to better describe the dynamic characteristics of the system. Firstly, a chaotic system is fused with the motion attitude data sequence, and then a suitable embedding dimension and time delay are randomly selected in the phase space of the chaotic system, and a set of initial trajectories are constructed. In the phase space, the system is evolved along this set of trajectories, and the rate of distance change between adjacent trajectories is calculated. Suppose these trajectories are $\mathrm{x_{1}(t),x_{2}(t),x_{3}(t)\ldots,x_{n}(t)}$, for adjacent points $x_{i}(t)$ and $x_{j}(t)$ in the trajectory, calculate the Euclidean distance between them. Then the rate of distance change between adjacent trajectories can be expressed as:

\begin{equation}\lambda(t)=\lim_{\delta_{t}\to0}\frac{1}{\delta_{t}}\ln(\frac{\left|x_{i}(t)-x_{j}(t)\right|}{\left|x_{i}(0)-x_{j}(0)\right|})\end{equation}

where $|\mathrm{x_{i}(t)-x_{j}(t)}|$ represents the distance between $x_{i}(t)$ and $x_{j}(t)$, and $|\mathrm{x_{i}(0)-x_{j}(0)}|$ represents their distance at the initial time, $\delta_{t}$ representing a small time interval. The average value of the rate of change of distance $\lambda(t)$ between adjacent trajectrajectory is calculated as the local Lyapunov index of each initial point in phase space. The maximum value of all local Lyapunov indices is taken as the Lyapunov index of the whole system, and the higher the value is, the worse the gait stability during walking.

\end{enumerate}

\subsection{\textbf{Statistical analysis}}
This study recruited 20 patients with lumbar disc issues (13 males, 7 females) and 15 healthy subjects (11 males, 4 females). The researchers recorded the height and weight data of each patient, as well as quantitative indicators of clinical scores. All gait parameters were analyzed in a single gait experiment on the affected side of patients with LDH, as well as on the left and right legs of healthy subjects. In order to determine if there are statistically significant differences in gait parameters between the healthy and affected sides, the Shapiro-Wilk normality test should be used initially to assess whether the gait parameters follow a normal distribution. When the significance value (p) is greater than 0.05, the parameters are considered to conform to the normal distribution. For parameters that do not have normality, the Mann-Whitney test method is used. Then, an independent sample T test was used to analyze the differences between gait parameters with normality on the healthy and affected sides of LDH patients. One-way analysis of variance was used to compare the differences in gait parameters between the left and right legs of LDH-affected sides and healthy subjects. Since 10 meters of gait parameters were collected for each subject in the walking experiment, the LDH healthy and affected side parameters were determined using the mean value within all gait cycles. The healthy gait parameters were calculated using the mean value of the gait parameters from both legs. A p-value less than 0.05 was considered statistically significant.

In order to validate the effectiveness of our multi-source data fusion method in assessing the gait ability of individuals with LDH lower limbs, we utilized the acceleration gait model developed by Rampp and Alexander \cite{35} to compute the spatiotemporal parameters of LDH gait. Specifically, we excluded the angle velocity from the calculation process for better accuracy. We plotted the radar plot of the gait spatio-temporal parameters calculated by the two methods. First, we performed standard normalization on all gait parameters, as shown in equation\ref{eq18}, where $\mu$ represents the mean of the parameters and $\sigma$ represents the standard deviation.

\begin{equation}X_{\mathrm{std}}=\frac{X-\mu}\sigma \label{eq18}\end{equation}

\begin{equation}OFF=\text{mean}(\sum_i|\text{LDH}_i-\text{H}|)\label{eq19}\end{equation}

For the standardized gait characteristics, we used healthy subjects as the baseline and calculated the average deviation values from the healthy and affected sides of LDH, relative to the baseline. This calculation is shown in equation\ref{eq19}, where ${LDH}_i$ and H represent the healthy and affected sides of LDH, respectively. The characteristic normalized value of the baseline determines the ability to identify LDH abnormal gait. A larger drift value indicates a better ability to identify LDH abnormal gait.

\subsection{\textbf{Classifier based on feature engineering}}
Random forest, multi-layer perceptron, and support vector machine classifiers were used to classify the gait parameters of the healthy and affected sides of LDH patients and healthy subjects. The purpose of the classifier was to validate the new gait feature model of LDH in differentiating between the healthy and affected sides of LDH. The validity of clinical differences between healthy subjects and the importance of assessing gait parameters in the clinical assessment of LDH are discussed.

The basic principle of the classifier is to distinguish gait abnormalities between healthy subjects and the affected side of LDH patients. In this approach, the gait parameters of healthy subjects are determined by taking the mean of the gait parameters of the left and right legs. The normal range of the gait pattern is defined by the gait parameters of the healthy group in the training dataset. Considering the potential intrinsic correlation between parameters, and the fact that the gait parameters of the healthy and affected side of LDH come from the same subject, it is necessary to establish a self-comparison model for LDH patients by conducting parameter screening. Therefore, a new set of classification parameters is obtained through the feature engineering method. For the new parameter set, the three types of parameter sequences for LDH healthy and affected sides, as well as healthy subjects, are reordered and then normalized using the max-min method to ensure they are on the same scale range. The data set is then divided into a training set, test set, and validation set to assess the generalization ability of the evaluation model. The data set is divided into 10 mutually exclusive subsets of similar sizes using ten-fold cross-validation, the partitioning process ensures that the consistency of the data distribution is maintained.

The Pearson correlation coefficient was calculated for all features and classification labels, and the feature importance was determined using the random forest method. For the random forest algorithm, the number of features was set to 9. The random forest was then traversed 100 times, and sampling was performed with replacement from the training set. This process was repeated n times to form a new sub-training set D. From this set, m features (where m $<$ 9) were randomly selected. A complete decision tree was then learned using the new training set D and the selected m features to obtain the random forest. The effectiveness of the new gait feature set was verified using SVM and MLP classifiers. We also perform the same classification on the gait parameter set calculated by the accelerometer data model to further verify the progress of our work. Finally, the accuracy, precision, and F1 scores are used to evaluate the classification effectiveness of the model.

\begin{equation}Accuracy=\frac{TP+TN}{TP+FN+FP+TN}\end{equation}
\begin{equation}Precision=\frac{TP}{TP+FP}\end{equation}
\begin{equation}F1=\frac{2TP}{2TP+FP+FN}\end{equation}

Among them, TP represents a true positive sample, which is a positive sample correctly predicted as positive by the model. FP represents a false positive sample, which is a negative sample incorrectly predicted as positive by the model. FN represents a false negative sample, which is a positive sample incorrectly predicted as negative by the model. TN represents a true negative sample, which is a negative sample correctly predicted as negative by the model.

\section{Result} \label{sec4}

\subsection{\textbf{Statistical analysis}}
We analyzed the biostatistical differences in gait patterns among four groups: LDH-affected side, LDH-affected side and healthy subjects, healthy subjects' left leg, and healthy subjects' right leg, as shown in table\ref{tab3}. In the comparison of the healthy and affected sides of LDH, the p-values for stride, pace, and single-step time in the spatiotemporal parameters of gait were 0.030, 0.011, and 0.002, respectively. Additionally, the p-value for stride time variability in the expansion index was 0.001. Among the kinematic parameters, the maximum bending angle and swing of the knee joint have p-values of 0.002 and 0.004, respectively, indicating that there are statistically significant differences in the self-contrast of LDH parameters. Additionally, step frequency, stride time, stance phase, swing time phase, and stride variability showed highly significant differences (p$<=$0.000), all of which were lower than the average level of healthy subjects. In comparing the affected side of LDH patients with healthy subjects, all gait parameters exhibited significant differences, indicating that pain significantly impairs the overall gait ability of LDH patients on the injured side. We also analyzed the gait parameters of the unaffected side of LDH patients and healthy subjects to investigate the influence of pain on the injured side on the unaffected side. We found significant differences in gait spatiotemporal parameters and expansibility indexes, which reflect the impact of pain on the affected side. The pain and numbness led to a deteriorated gait pattern, resulting in decreased regularity and stability while walking. Additionally, this confirmed the clinical diagnosis of the healthy and affected side. The kinematic parameters, specifically the maximum knee flexion angle and swing angle, had p-values of 0.460 and 0.107, respectively. This lack of significant difference is attributed to variations in age and walking conditions. In order to confirm that the disparity between the healthy and affected sides of LDH was not coincidental, we examined the gait parameters of both the left and right legs of all healthy subjects. The analysis revealed no significant differences in any of the subjects.

\begin{table*}[htpb]%调节图片位置，h：浮动；t：顶部；b:底部；p：当前位置
	\caption{Statistical analysis data of three types of gait pattern parameters.}
	\label{tab3}  
	\begin{tabular}{ccc ccc ccc}%表格中的数据居中，c的个数为表格的列数
		\hline\noalign{\smallskip}	
		&  &Gait pattern  &  &  &  & p value &\\
		\noalign{\smallskip}\hline\noalign{\smallskip}
            \noalign{\smallskip}
		   Features & LDHH & LDHE & H  & HE & H-H & E-H & LR\\
            \noalign{\smallskip}\hline\noalign{\smallskip}
		  SF&1.48($\pm$0.74) & 1.08($\pm$0.78) & 1.96($\pm$0.40)  &0.000 &0.000 &0.000 &0.889\\
            SL & 0.94($\pm$0.98)  &	0.73($\pm$0.74)   &1.41($\pm$0.56)	&0.030	&0.000	&0.000	&0.164 \\
            SS& 0.50($\pm$0.93)	&0.35($\pm$0.72)		&0.85($\pm$0.56)	&0.011	&0.000	&0.000	&0.322\\
             STT & 0.98($\pm$0.91)	&1.23($\pm$1.21)	&0.76($\pm$0.40)	&0.002	&0.008	&0.000	&0.854\\
             ST & 1.31($\pm$0.61)	&1.74($\pm$0.82)	&1.02($\pm$0.26)	&0.000	&0.020	&0.000	&0.823\\
            STP & 0.57($\pm$0.55)&0.38($\pm$1.06)	&0.62($\pm$0.28)	&0.000	&0.000	&0.000	&0.836\\
            SWP & 0.43($\pm$0.55)	&0.62($\pm$1.06)	&0.38($\pm$0.28)	&0.000	&0.000	&0.000	&0.797\\
            SV & 0.12($\pm$0.88)	&0.24($\pm$0.84)	&0.06($\pm$0.57)	&0.000	&0.004	&0.000	&0.289\\
            STV & 0.15($\pm$0.79)	&0.25($\pm$0.82)	&0.04($\pm$0.79)	&0.001	&0.000	&0.000	&0.936\\
            STA & 0.24($\pm$0.99)	&0.26($\pm$0.84)   &0.22($\pm$0.78)	&0.033	&0.014	&0.000	&0.878\\
             WQK & 40.72($\pm$1.26)&23.16($\pm$0.65)	&38.71($\pm$0.76)	&0.002	&0.460	&0.000	&0.903\\
            BDK & 26.64($\pm$0.87)	&40.32($\pm$1.02)	&23.97($\pm$0.74)	&0.004	&0.107	&0.000	&0.187\\
		\hline\noalign{\smallskip}  
	\end{tabular} 
 {\footnotesize \small $\bullet$ Gait parameter values of LDH-affected side and healthy subjects, all values are presented in the form of mean and standard deviation. HE: Self-comparison of healthy and affected sides, LR: Self-contrast of healthy subjects, H-H: Comparison of LDH healthy side and healthy side, E-H: Comparison of LDH affected side and healthy side, p-value (p$<$0.05 indicates a significant difference, p$=$ 0.000 indicates a strongly significant difference).}
\end{table*}

\begin{figure}
    \centering
    \includegraphics[width=1\linewidth]{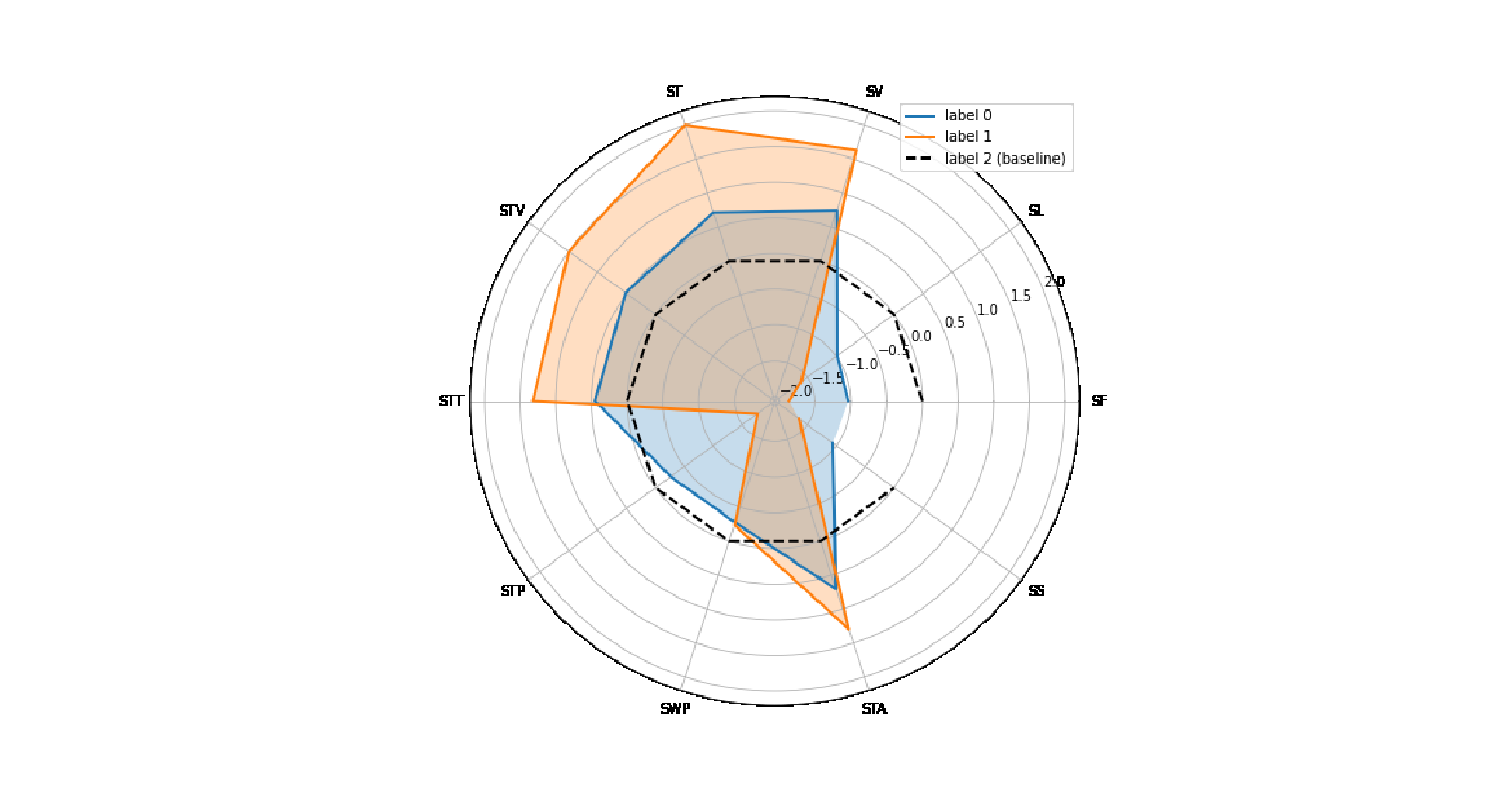}
    \caption{Fusion gait feature radar chart, where label 0 is the healthy side of LDH, 1 is the affected side of LDH, and 2 is the healthy baseline.}
    \label{fig5}
\end{figure}

\begin{figure}
    \centering
    \includegraphics[width=1\linewidth]{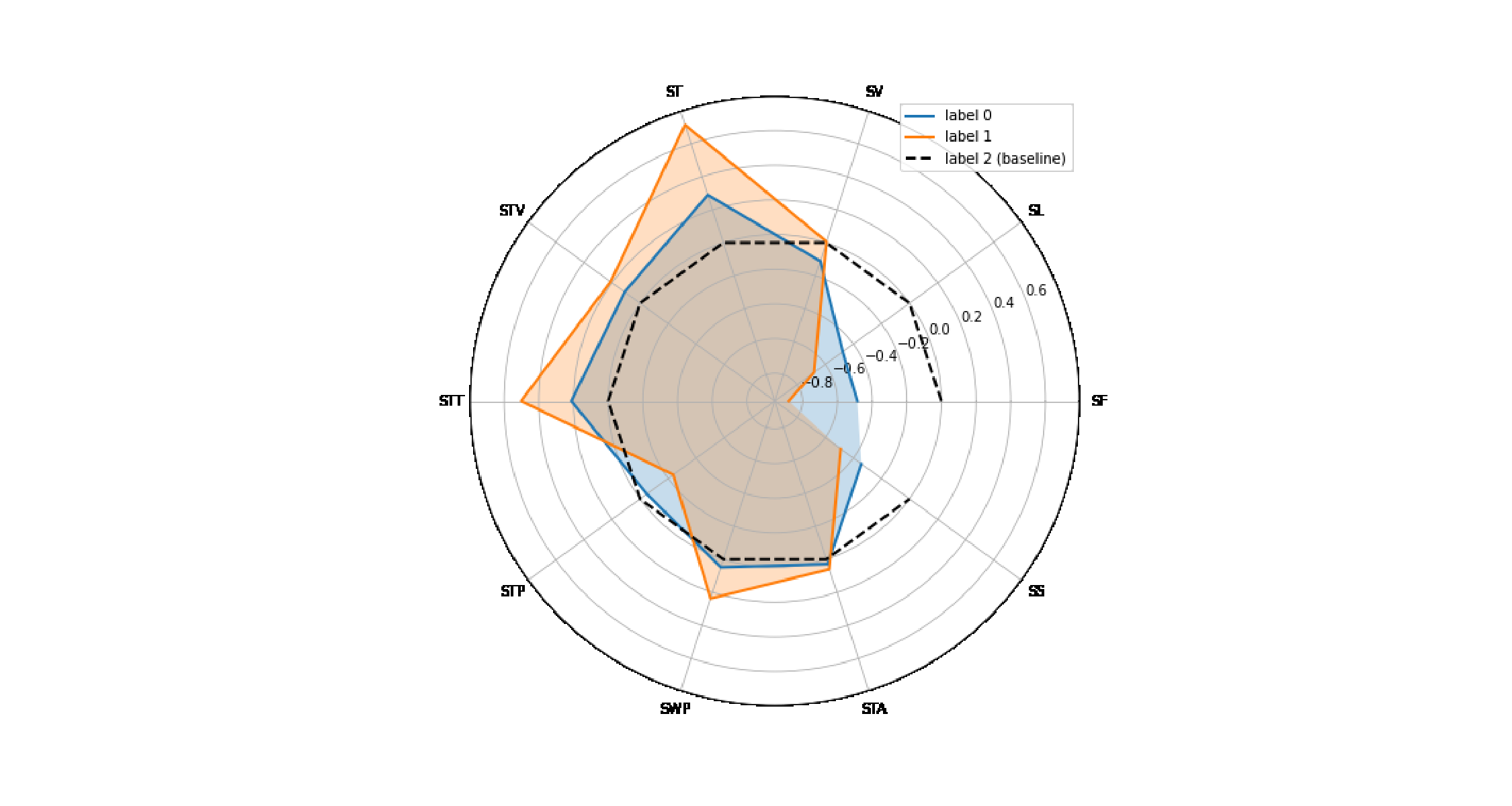}
    \caption{ACC model gait characteristic radar chart}
    \label{fig6}
\end{figure}

We utilized the acceleration model to compute the spatiotemporal parameters of gait and determined the average deviation value of both the healthy and affected sides of LDH, which was compared to the healthy baseline. Patients with LDH experience severe motor function deterioration in the lower limbs, and numbness on the affected side can result in reduced gait ability on the unaffected side while walking \cite{Alakokko2002GeneticRF}. As shown in table\ref{tab4}, the average drift value of the healthy side of LDH(OFF0) obtained from the acceleration model is 0.22, while the average drift value of the affected side of LDH(OFF1) is 0.40. The average drift values of the healthy and affected sides of LDH, calculated by our fusion feature model, are 0.67 and 1.49. Higher than the ACC model, our feature set can effectively characterize the degradation of lower limb motor function in LDH patients and clearly distinguish between the healthy and affected sides. Figures \ref{fig5} and \ref{fig6} display radar distribution diagrams illustrating the gait parameters of both the healthy and affected sides of individuals with LDH, as well as healthy subjects. The mean parameters of the healthy subjects serve as the baseline, and the LDH healthy and affected sides, calculated using two characteristic models, are plotted in comparison to the baseline. The average difference value between. For our fused gait model, the deviation trends of STT, ST, SL, SF, STP, and SS are consistent with the ACC model.
\begin{table}
\centering
\caption{Comparison of ACC model and radar plot of our model.}
\begin{tabular*}{0.5\textwidth}{@{\extracolsep{\fill}}l*{2}{l}}
\toprule
Data & OFF0 & OFF1 \\
\midrule
ACC & 0.22 & 0.40 \\
Ours & 0.67 & 1.49 \\
\bottomrule
\end{tabular*}
\label{tab4}
\end{table}
Additionally, the variability index and gait stability characterization effects are significantly improved compared to the ACC model. In addition, the ST and SL parameters on the healthy side of LDH have the largest offset values among all parameters. This suggests that the decline in gait ability on the affected side also affects the ability of the healthy side. For the stance and swing phases, the fusion gait model highlights the decreased gait ability of the affected side of LDH. Additionally, the trend of the healthy side of LDH being affected by the affected side is consistent with the ACC model.

\subsection{\textbf{Classifier based on feature engineering}}
To investigate the factors that influence the gait ability of patients with LDH, we assessed the significance of various gait characteristics and their Pearson correlation coefficient with the three gait-like patterns, the results are presented in table\ref{tab5}. We employed the random forest feature engineering method to identify a set of 9 features for classification purposes. For feature correlation, SF, SL, STP, SS, and WQK showed a positive correlation with gait mode. Among these, SL and SS had the highest correlation with gait mode. On the other hand, SV, ST, and STT showed a negative correlation with gait mode. Correlation, among which SV and ST have the highest correlation values. For feature importance, we selected the top 9 important features as the parameter set. Among these features, SF has the highest importance in random forest classification at 0.26\%, while STV and WQK have the lowest importance at 0.05\%. This indicates that SF has the highest recognition ability among the three gait-like patterns.

\begin{figure}
    \centering
    \includegraphics[width=0.6\linewidth]{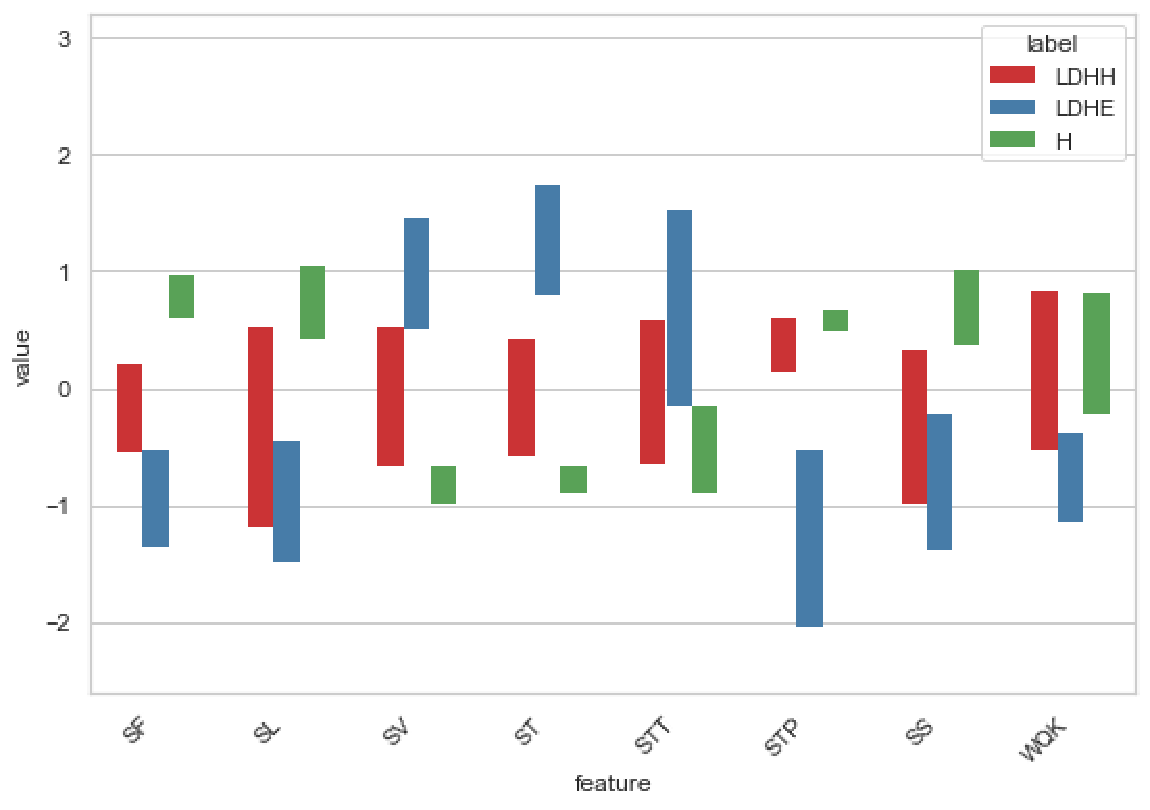}
    \caption{Gait parameter distribution box plot.}
    \label{fig7}
\end{figure}

Figure\ref{fig7} shows the box plot of the gait parameter model after feature screening. It illustrates the standard normalized distribution of gait parameters on the healthy side, affected side, and healthy subjects with LDH, respectively. The longer box plot represents that the subject group has unstable gait ability. Among them, the gait parameters of healthy subjects are used as the benchmark. The overall ability of patients with LDH in terms of gait spatiotemporal parameters and expansibility indicators is lower than that of healthy subjects. We found that patients with LDH (lumbar disc herniation) had a more stable overall fluctuation of the maximum knee flexion angle compared to healthy subjects. This suggests that pain restricts the range of motion of the knee joint during exercise. In the self-comparison of LDH patients, the amplitudes of stride length, stride variability, stride time, stride time variability, stride speed, and maximum knee joint flexion angle on the unaffected side were higher than those on the affected side. This indicates that the damage to the affected side has caused worse stability on the unaffected side compared to the original. Additionally, the fluctuation in the value range of overall gait parameters is smaller in healthy subjects compared to LDH patients. This suggests that the overall walking ability of LDH patients is worse than that of healthy subjects. The parameter box of healthy subjects. The graph represents the overall gait baseline and serves as a reference value for clinical comparison.

\begin{table}
  \centering
  \caption{Feature relevance and importance.}
  \begin{tabular*}{0.5\textwidth}{@{\extracolsep{\fill}}l*{2}{l}}
    \toprule
    Features & Correlation & Importance  \\
    \midrule
    SF & 0.51 & 0.26\% \\
    SL & 0.56 & 0.09\% \\
    SV & -0.38 & 0.04\% \\
    ST & -0.38 & 0.15\% \\
    STV & -0.28 & 0.05\% \\
    STT & -0.21 & 0.09\% \\
    STP & 0.19 & 0.11\% \\
    SS & 0.56 & 0.14\% \\
    WQK & 0.33 & 0.05\% \\
    \bottomrule
  \end{tabular*}
  \label{tab5}
\end{table}

The accuracy, precision, and F1 scores of the three proposed classifiers in different classification tasks of the two models are shown in tables \ref{tab6} and \ref{tab7}. In order to assess the effectiveness of the features in distinguishing between the healthy and affected sides of LDH, we conducted an analysis on both sides of LDH patients. Gait patterns of healthy subjects were classified. The parameter set, after feature screening, showed good classification results on the three classifiers. Among them, the SVM classifier achieved the highest accuracy of 92.86\%. Additionally, the precision and F1 score for this classifier were 94.64\% and 92.93\% respectively. The ACC model achieved the highest accuracy of 90\% on the RF classifier. Additionally, all the indicators of the fusion feature model are higher than the highest indicators of the ACC model. We also classified individual LDH patients and healthy subjects.The MLP classifier achieved the optimal classification effect, with accuracy, precision, and F1 score of 95.50\%, 97.75\%, and 95.86\% respectively.These scores are higher than those achieved by the SVM classifier, indicating that the MLP model has the highest classification effect. It has been verified that the new gait parameter model can assist in the clinical diagnosis of patients with LDH and distinguish between the healthy and affected sides. The classifier, after feature screening, has a better classification effect and can more accurately reflect the biological differences between the healthy side, affected side, and healthy subjects with LDH.
\begin{table}
\centering
\caption{Average accuracy, precision and f1 score of three classifications on the affected side of LDH patients and healthy subjects.}
\begin{tabular*}{0.5\textwidth}{@{\extracolsep{\fill}}l*{2}{ll}}
\toprule
Model & Accuracy & precision  & F1 \\
\midrule
RF-ACC & 90.00\% & 92.00\% &89.80\%\\
RF-ours & 86.00\% & 89.79\% &84.91\%\\
MLP-ours & 85.71\% & 88.57\% &82.93\%\\
\textbf{SVM-ours} & \textbf{92.86\%} & \textbf{94.64\%}&\textbf{92.93\%}\\
\bottomrule
\end{tabular*}
\label{tab6}
\end{table}

\begin{table}
\centering
\caption{Average accuracy, precision and f1 score for binary classification in LDH patients and healthy subjects.}
\begin{tabular*}{0.5\textwidth}{@{\extracolsep{\fill}}l*{2}{ll}}
\toprule
Model & Accuracy & Precision  & F1 \\
\midrule
SVM-ACC & 92.50\% & 93.33\% & 93.33\%\\
RF-ours & 91.00\% & 97.33\% & 93.88\%\\
SVM-ours & 90.50\% & 94.42\% & 90.52\%\\
\textbf{MLP-ours} & \textbf{95.50\%} & \textbf{97.75\%} & \textbf{95.86\%}\\
\bottomrule
\end{tabular*}
\label{tab7}
\end{table}

\section{Discussion}\label{sec5}
This article proposes a new method for the clinical evaluation of lumbar disc herniation based on the fusion of multi-source data from IMUs . Acceleration data and angular velocity data were collected using an IMU attached to the calf. Two IMUs were used to collect gait information. This method is lightweight, low-cost, and reduces the interference of other human activities on gait. Our data is collected under strict professional physician guidance during the labeling process. In the literature \cite{23}, Çelik et al. summarized the method of multi-source data fusion for clinical gait recognition. They believed that acceleration and angular velocity data were the most suitable data to estimate gait kinematics. Mitchell et al. proposed the use of an enhanced complementary filter to combine acceleration and angular velocity data for human activity recognition [34]. Previous studies have extensively shown the significance of utilizing multi-source data in gait recognition tasks, which aligns with our research focus. Current LDH disease gait recognition models focus on mining information from single-source acceleration data \cite{Kuligowski2021LumbopelvicBI,3Lee2021AssociationBP,11,12,13,14,36}. Specifically, the adaptive Kalman data fusion algorithm is used to combine the Euler angle calculated from acceleration and angular velocity. The attitude angle of the motion process is obtained through cumulative integration. Additionally, the peak detection algorithm is utilized to divide the gait phase and determine the gait space-time. The study focuses on evaluating the stability of the motion state by considering expandability and kinematic parameters, analyzing the timing of attitude angles, and utilizing the Eckmann algorithm to calculate the maximum Lyapunov index. Taking multi-source sensor data into consideration reduces errors in calculating gait parameters.

We have developed a new framework for clinically identifying multiple types of gait parameters between the healthy and affected sides of individuals with LDH, as well as healthy subjects. This framework includes 12 parameters, such as gait spatiotemporal parameters, kinematic parameters, and expansibility indicators. A total of 32 subjects were recruited. The gait patterns of the subjects were divided into three groups: LDH-unaffected side, LDH-affected side, and healthy subjects. As shown in table\ref{tab3}, the statistical analysis revealed significant differences in walking cadence, stride time, stance phase, swing phase, and stride variability between the healthy and affected side of LDH. This provides a basis for clinical judgment of patients with LDH. Severity and movement profile provide strong indicators of differentiation from healthy individuals. Low back pain and numbness in the lower limbs are the most prominent clinical symptoms in patients with LDH. Lower limb numbness will lead to a reduction in the range of motion of the affected side's knee joint, but it does not affect the movement of the healthy side's knee joint. In terms of overall gait performance, the affected side of LDH was worse than the healthy side and healthy subjects, while the healthy side of LDH was also worse than healthy subjects. Since factors such as age, gender, and weight can influence the measurement of gait parameters \cite{15,37}, it is important to note that the control group selected for this study consisted of individuals with varying ages, weights, and walking conditions, including both healthy subjects and LDH patients. This inconsistency in the control group may introduce errors and affect the significance of the observed differences during comparison.

In order to validate the effectiveness of the proposed gait parameter framework, we developed a machine learning model that utilizes feature engineering to classify gait parameters. RF was used to calculate the feature importance and Pearson correlation coefficient to determine the correlation between the three categories. Gait parameters are highly correlated with the gait pattern. As shown in Tables 5 and 6, we utilized various indicators from different computing theories, namely accuracy, precision, and recall, for the gait pattern classification task. Accuracy measures the overall performance of the classification algorithm, precision measures the correctness of the prediction algorithm, and the F1 score reflects the algorithm's sensitivity to imbalanced data. We utilized machine learning methods, specifically Support Vector Machines (SVM) and Random Forest (RF), as well as deep learning method, Multi-Layer Perceptron (MLP), to classify three different gait patterns. The SVM model has better generalization ability and is more effective in identifying the affected side of LDH and healthy subjects. The effect is highest in the task. MLP demonstrates the best performance in identifying the overall gait of LDH patients and healthy subjects. Both classifiers show good classification effects in the task of classifying imbalanced data. It has been verified that our gait model can effectively distinguish between the healthy and affected sides of LDH in both healthy subjects and patients. This provides clinicians with an effective gait recognition method to assist in diagnosing the affected sides of LDH.

Our current work mainly focuses on feature engineering methods to assess the effectiveness of biometrics. However, this work still has some limitations. During the data collection process, the collection environment is clinically performed in a hospital, and the subjects walk in a straight line. As a result, the gait changes during the turning process are not taken into account. In addition, it is difficult to control the balance between the age, height, and other data of healthy subjects and the LDH patient group. Future work can focus on identifying different walking patterns of patients with LDH, performing data augmentation or expanding the dataset, and combining multimodal fusion methods to design deep learning models that can better explore the nonlinear relationships of pathological indicators. Additionally, research can be conducted on real-life scenarios. Experiments to improve the comprehensiveness and efficiency of auxiliary clinical diagnosis.

\section{Conclusion}\label{sec6}
In this article, we focus on establishing a gait parameter model using IMU multi-source data fusion to identify abnormal gait in individuals with LDH. To achieve this, we considered fusing acceleration and angular velocity data to segment the corresponding gait phases. This allowed us to obtain spatiotemporal parameters, kinematic parameters, and extensibility indicators. Segmenting gait phases is challenging due to signal acquisition and the utilization of different types of data. Therefore, we propose an adaptive Kalman data fusion algorithm to fuse the attitude angle and segment the gait phase by converting acceleration and angular velocity into Euler angles. Statistical analysis of gait parameters and gait pattern recognition using different classifiers on the collected IMU multi-source datasets demonstrated the advantages of this method.

\section*{Data availability statement}
The data cannot be made publicly available upon publication because they contain sensitive personal information. The data that support the findings of this study are available upon reasonable request from the authors.

\section*{Acknowledgment}
This work was funded by Changzhou Science and Technology Planning Project (Grant No.: CJ20235026) and Jiangsu Province Postgraduate Research Practice Innovation Plan Project (Project No.: KYCX23\_3070). 

\section*{ORCID iDs}
\begin{flushleft}
Yongsong Wang: \href{https://orcid.org/0009-0000-9545-4416}{\textcolor{green}{https://orcid.org/0009-0000-9545-4416}}

Lin Chen: \href{https://orcid.org/0009-0008-6486-7988}{\textcolor{green}{https://orcid.org/0009-0008-6486-7988}}
\end{flushleft}

\end{document}